\documentclass[a4paper]{jpconf}
\usepackage{iopams}
\usepackage{citesort}
\usepackage{graphicx}
\begin{document}
\title{Precise Payload Delivery via Unmanned Aerial Vehicles: An Approach Using Object Detection Algorithms}

\author{Aditya Vadduri$^1$, Anagh Benjwal$^2$, Abhishek Pai$^3$, Elkan Quadros$^1$,  Aniruddh Kammar$^3$ and Prajwal Uday$^3$}

\address{$^1$B.Tech Mechatronics Engineering, Manipal Institute of Technology, Manipal Academy of Higher Education, Manipal, Karnataka, India-576104}
\address{$^2$B.Tech Electronics \& Communication Engineering, Manipal Institute of Technology, Manipal Academy of Higher Education, Manipal, Karnataka, India-576104}
\address{$^3$B.Tech Mechanical Engineering, Manipal Institute of Technology, Manipal Academy of Higher Education, Manipal, Karnataka, India-576104}

\ead{vadduriaditya@gmail.com, benjwal.anagh@gmail.com, abhivpai@gmail.com, johnelkanquadros@gmail.com, aniruddh.kammar@gmail.com, uprajwal20@gmail.com}

\begin{abstract}
Recent years have seen tremendous advancements in the area of autonomous payload delivery via unmanned aerial vehicles, or drones. However, most of these works involve delivering the payload at a predetermined location using its GPS coordinates. By relying on GPS coordinates for navigation, the precision of payload delivery is restricted to the accuracy of the GPS network and the availability and strength of the GPS connection, which may be severely restricted by the weather condition at the time and place of operation. In this work we describe the development of a micro-class UAV and propose a novel navigation method that improves the accuracy of conventional navigation methods by incorporating a deep-learning-based computer vision approach to identify and precisely align the UAV with a target marked at the payload delivery position. This proposed method achieves a 500\% increase in average horizontal precision over conventional GPS-based approaches.

\end{abstract}

\section{Introduction}

\subsection{Background}
In 2018, the Indian Ministry of Civil Aviation issued new guidelines for the use of drones in commercial applications, leading to a significant expansion in the drone market. By 2026, the drone market in India is projected to reach a valuation of \$46 billion \cite{michelindia}.

There is a growing interest in using unmanned aerial vehicles (UAVs) for delivery applications, including medical relief, merchandise, and food packages. The package delivery industry presents a significant opportunity for the implementation of UAV technology. E-commerce giants like Amazon have started phased trials \cite{tarasov_2022} to test the viability of package delivery using  UAVs to increase their supply-chain efficiency. Studies conducted suggest that upto ~30\% of EU residents \cite{aurambout2019last} will be able to benefit from UAV deliveries while also keeping factors like traffic and emissions in check.

Conventional drone-based delivery systems rely on Global Positioning System (GPS) or Global Navigation Satellite System (GNSS) based navigation as shown in \cite{vehiclerouting,dronepsych,dronesecurity}. The GPS coordinates of the intended delivery location and the real-time GPS coordinates of the drone during the mission are used to guide the UAV along the intended trajectory for delivering the payload. This method has been proven successful for applications such as delivery of medical and emergency supplies \cite{dronedefib,medicinedel}, product delivery \cite{productdelivery}, and other such outdoor navigation scenarios.

\subsection{Problem Overview}

UAVs use circularly polarized upward-facing antennas to establish a GPS connection; The ArduCopter autopilot system has a built-in requirement of at least 8 satellites to allow fully autonomous flight. Conventional GPS modules used in UAVs tend to have a horizontal position accuracy of 3 to 5 meters \cite{rychlicki2020analysis}.

Increased positional accuracy can be achieved by equipping the drone with additional sensors, such as a rangefinder, Light Detection and Ranging (LiDAR), or optical flow sensor, for improved perception as shown in \cite{campus2021autonomous,8324394,8999672}. However, these sensors are expensive and not readily available.

In many commercial applications, particularly for navigation in urban or indoor environments, a horizontal positional accuracy of less than 1 meter is required, which is impossible with a  conventional GPS module. Moreover, the accuracy and strength of the signal is significantly impacted by up to 70dB \cite{inproceedings} due to poor weather conditions  leading to a high path loss, which makes it unreliable to use in many cases.

It is essential to address these issues in order to facilitate the growth and integration of UAV technology into our daily lives.

\subsection{Proposed Solution}

In this work, a robust approach to autonomous drone delivery, which improves upon traditional GPS-reliant methods, is presented. The proposed method, which utilizes a two-part system incorporating GPS-based autonomous navigation and a deep-learning-based object detection algorithm, requires the presence of a target at the intended delivery location. The size of the target used in the experiments has been depicted in Figure 1.


\begin{figure}[h!]
\centering
\includegraphics[width=2in]{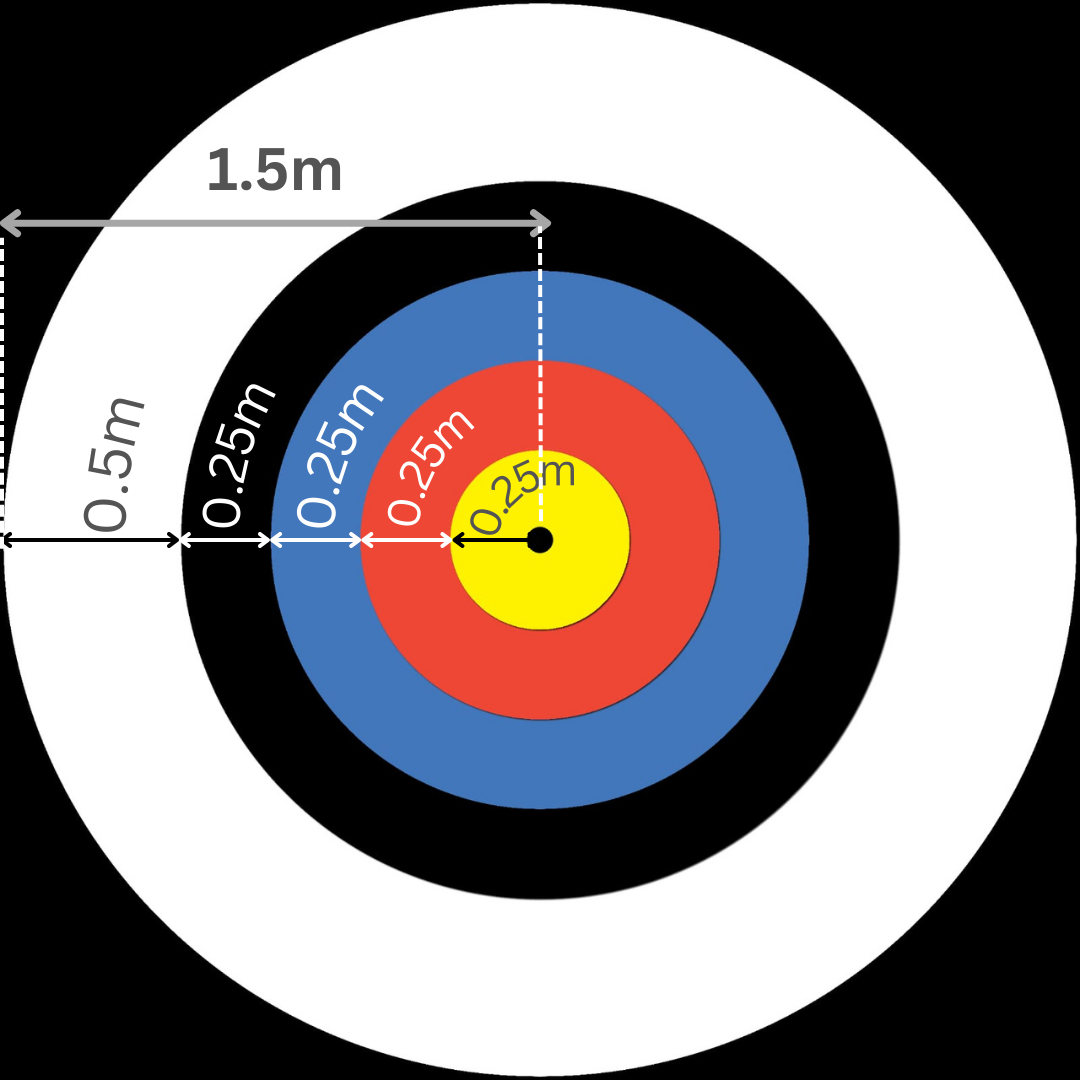} \begin{minipage}[b]{14pc}\caption{\label{label}Target used to mark payload delivery location}
\end{minipage}
\end{figure}

The GPS-based navigation system guides the UAV to the vicinity of the delivery location. Once the UAV reaches this location, the object detection algorithm takes over. This algorithm processes the video feed from the UAV's camera using a convolutional neural network to detect the target's location. The UAV then adjusts its position to ensure that the target is aligned with the center of the camera frame.

The practicality of the proposed method has been demonstrated by building a micro-class quadcopter UAV with an all-up weight of 1.95 kg and a flight time of approximately 8 minutes. The details of the structure of the drone and the hardware components used have been outlined below. Our results show that the proposed navigation algorithm improves accuracy and precision in payload delivery compared to traditional GPS-only methods.

\section{Methodology}

\subsection{Design Overview}

The materials used in the manufacturing of components of the UAV were finalized after being subjected to comprehensive bending, tensile, and torsional tests. The use of additive manufacturing processes, specifically 3D printing, allowed for the efficient and precise fabrication of the structural components of the UAV. In 3D printing, a digital model is used to build up the object layer by layer, using materials such as plastics, metals, or composites, until the desired final shape is achieved. 

The wheelbase of the UAV consists of two 1mm carbon fiber sheets, which are held in place by structural components that are fabricated from  Acrylonitrile Butadiene Styrene (ABS) and Polyacetic acid (PLA) plastic. The rotor arms and landing gears of the UAV are fashioned from pultruded carbon fiber rods with a cross-section of $10mm \times 10mm$ and $8mm \times 8mm$, respectively. The payload bay is also fabricated from PLA plastic and employs a servo-actuated mechanism; it is designed to accommodate a moderate-sized payload of dimensions $10cm \times 5cm \times 5cm$. 

\begin{figure}[h!]
\begin{minipage}{0.48\textwidth}
\centering
\includegraphics[width=2.5in]{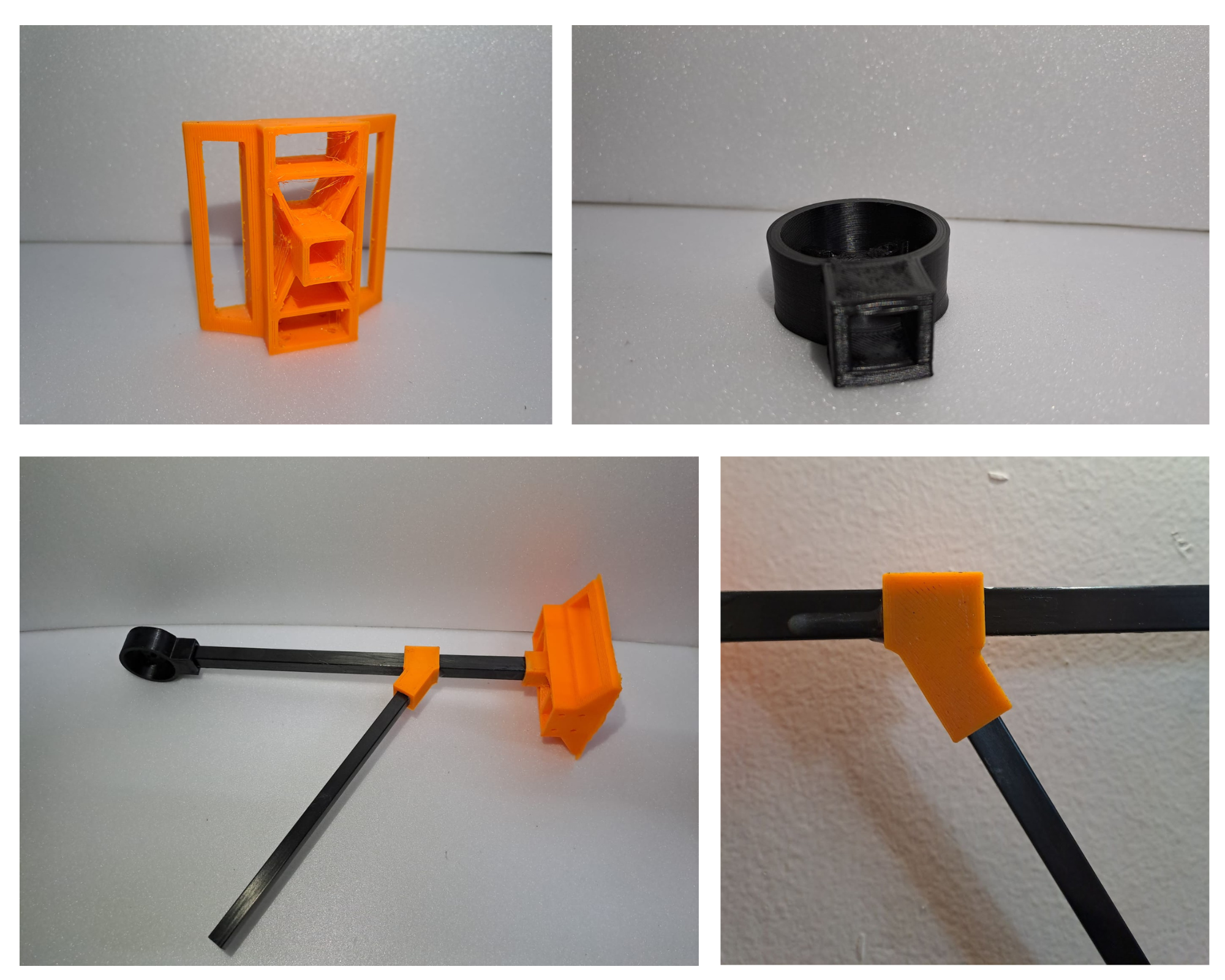} 
\end{minipage}%
\begin{minipage}{0.48\textwidth}
\centering
\includegraphics[width=2in]{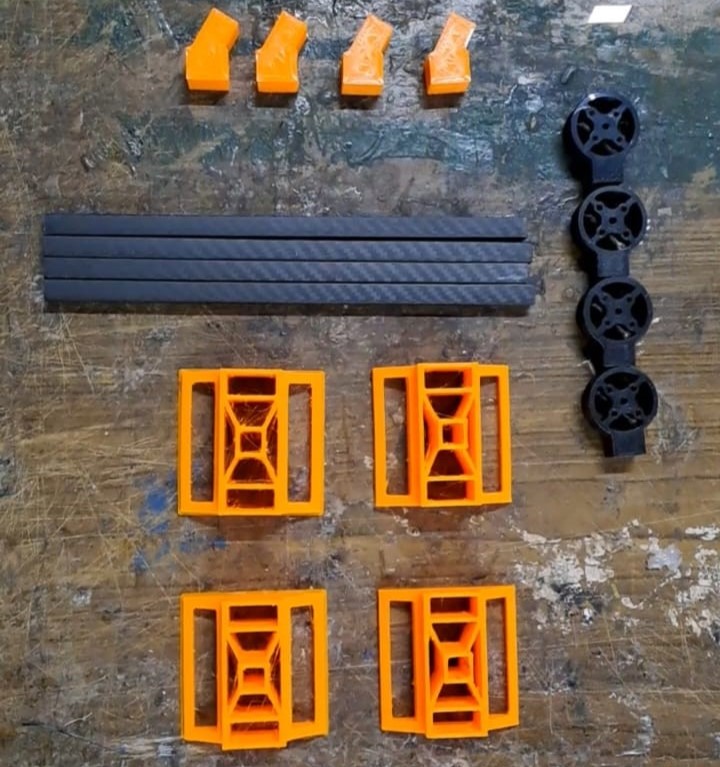} 
\end{minipage}
\caption{Manufactured Components for UAV}

\end{figure}

\subsection{Hardware Overview}

The UAV is built in a quadcopter configuration, i.e. with 4 brushless DC motors that are responsible for the propulsion of the UAV. The flight controller uses either manual inputs from a transmitter (received via an on-board radio receiver) or positional commands sent by the on-board flight computer to calculate the necessary Pulse Width Modulation (PWM) values for each motor. These values are sent to the 4-in-1 Electronic Speed Controller (ESC), which then adjusts and controls the speed of the corresponding motor.

\begin{figure}[h!]
\centering
\includegraphics[width=6.2in]{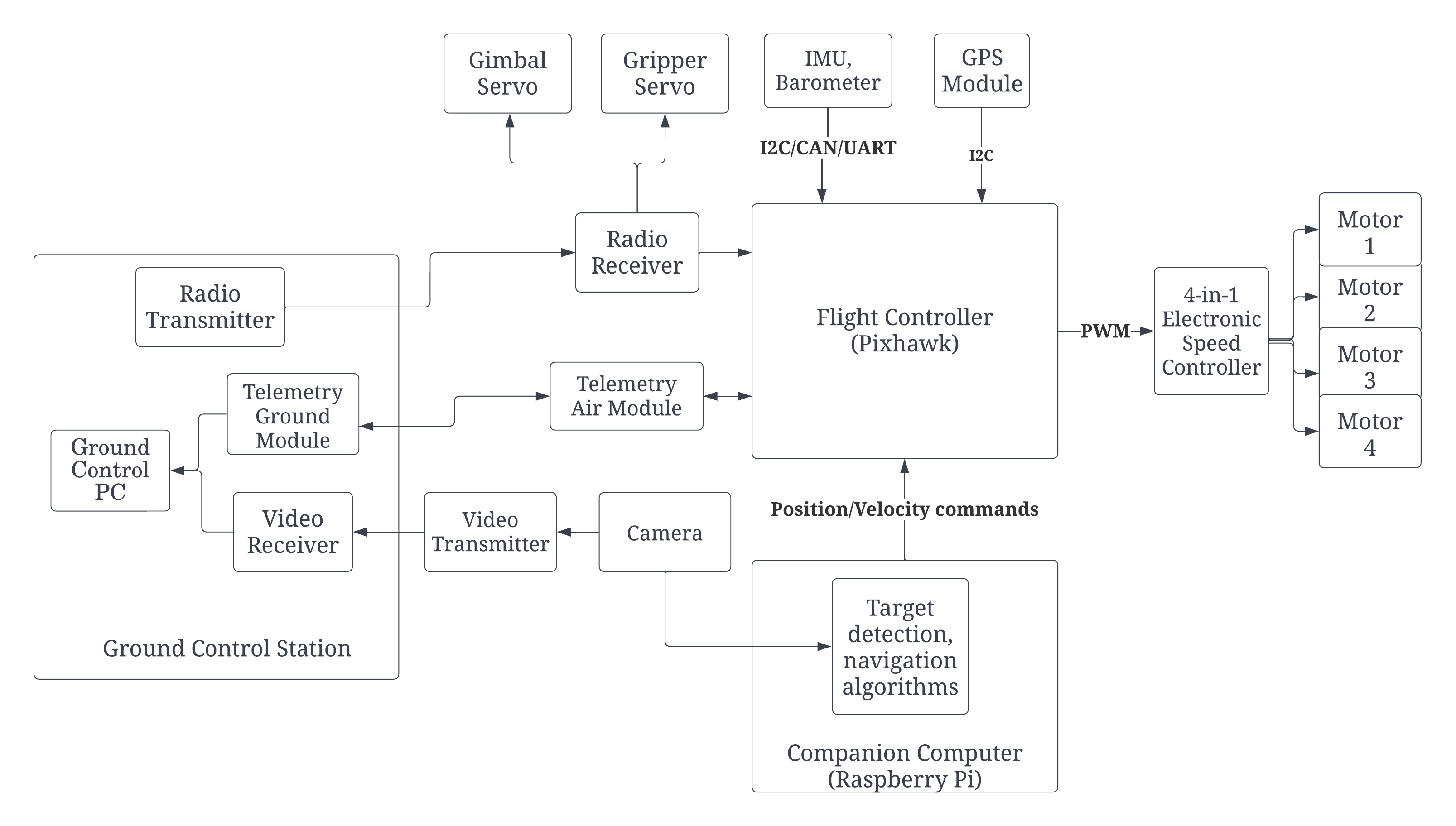} 
\caption{Overview of UAV Operation}
\end{figure}

\subsection{Propulsion System}

The T-Motors AT2317 motors, paired with 9$\times$4.5 inch propellers were selected for the UAV, This combination provided a maximum total thrust of $4.3 kgf$. The T-motor F45 4-in-1 ESC was chosen to control the motor-propeller combination. A 6200mAh 3S Lithium Polymer (LiPo) battery was selected for the UAV to achieve an average flight time of 8 minutes while keeping the all-up-weight of the UAV nearly 2 kilograms. The performance parameters of the UAV were calculated using equations from \cite{uavcalc,equationsuav} and have been listed in Table 1.

\begin{table}[h!]
\caption{\label{ex}UAV Performance Parameters}
\begin{center}
\begin{tabular}{lllllll}
\br
\begin{tabular}[c]{@{}l@{}}Propeller dimensions - 9 $\times$ 4.5\\ Total UAV Weight - 1953g\end{tabular} & \multicolumn{6}{c}{Throttle Percent}               \\ \hline
                                                                                                & 16.6\% & 33.3\% & 50\%  & 66.6\% & 83.3\% & 100\%  \\ \hline
Current draw per Motor($A$)                                                                       & 1.39   & 4.26   & 7.52  & 13.09  & 20.75  & 30.27  \\
Power drawn per Motor ($W$)                                                                       & 17     & 49.7   & 89.1  & 141    & 222.3  & 311    \\
Thrust per Motor ($gf$)                                                                           & 122    & 278    & 449   & 670    & 938    & 1077   \\
Efficiency of Motor ($gf/W$)                                                                      & 7.17   & 5.59   & 4.84  & 4.75   & 4.19   & 3.46   \\
Total Current draw of UAV ($A$)                                                                   & 6.66   & 18.14  & 31.18 & 53.46  & 84.1   & 122.18 \\
Total Power drawn by UAV ($W$)                                                                    & 73.5   & 204.3  & 361.9 & 569.5  & 894.7  & 1249.5 \\
Total Thrust ($gf$)                                                                               & 488    & 1112   & 1796  & 2680   & 3752   & 4308   \\
Thrust-to-Weight Ratio                                                                          & 0.24   & 0.56   & 0.91  & 1.37   & 1.92   & 2.21\\  
Tilt Angle ($^\circ$)                                                                                                   & - & - & - & 43.21       & 58.63      & 63.04     \\
Takeoff Velocity ($m/s$)                                                                                                 & - & - & - & 8.29        & 9.81       & 10.52     \\
Horizontal Velocity ($m/s$)                                                                                              & - & - & - & 11.28       & 15.94      & 18.08     \\
Flight Time (minutes)                                                                                                  & - & - & - & 6.95        & 4.42       & 3.044     \\
Range (kilometers)                                                                                                     & - & - & - & 4.70        & 4.22       & 3.30  \\ \br 
\end{tabular}
\end{center}
\end{table}

\subsection{Communication System}

The UAV maintains constant communication with the Ground Control Station (GCS) through a telemetry module that ensures real-time transmission of flight data from the UAV to the GCS. 
 
In addition to these communication links, a secondary connection was established between the GCS and the companion computer via a Long Term Evolution (LTE) module, allowing for  remote access to the onboard computer on the UAV through a Secure Shell (SSH) connection. The range and functionality of all established avionics and connections of the system were tested and verified to a minimum range of 5 kilometers manually.

\subsection{Autopilot and On-board Computation System}

The UAV is equipped with a Pixhawk 2.4.8 flight controller and uses the ArduCopter autopilot firmware \cite{missionplanner}. This flight controller was selected due to the in-built 9-axis Inertial Measurement Unit (IMU), gyroscope, barometer, accelerometer, and its connectivity options that include multiple Universal Asynchronous Receiver Transmitter (UART), Inter Integrated Circuit (I2C), and Serial Peripheral Interface (SPI) ports, as well as a 32-bit redundant processor that ensures robustness and high mission success rates.

The Raspberry Pi 4 was selected as the companion computer that is used to execute the autonomous navigation and target detection algorithms developed. This micro-computer is equipped with a 64-bit quad-core processor with a clock speed of 2.4GHz and 8GB of memory \cite{raspberrypi}. This ensures that the UAV has adequate computational capabilities to run the developed algorithms. 

Additionally, the UAV is equipped with a PiCamera V2 camera module, which is compatible with the Raspberry Pi 4 and is capable of capturing videos at 30 frames-per-second at a resolution of 1080p. The video feed of this camera serves as the input to the developed target detection algorithm.

\begin{figure}[h]
\centering
\includegraphics[width=3in]{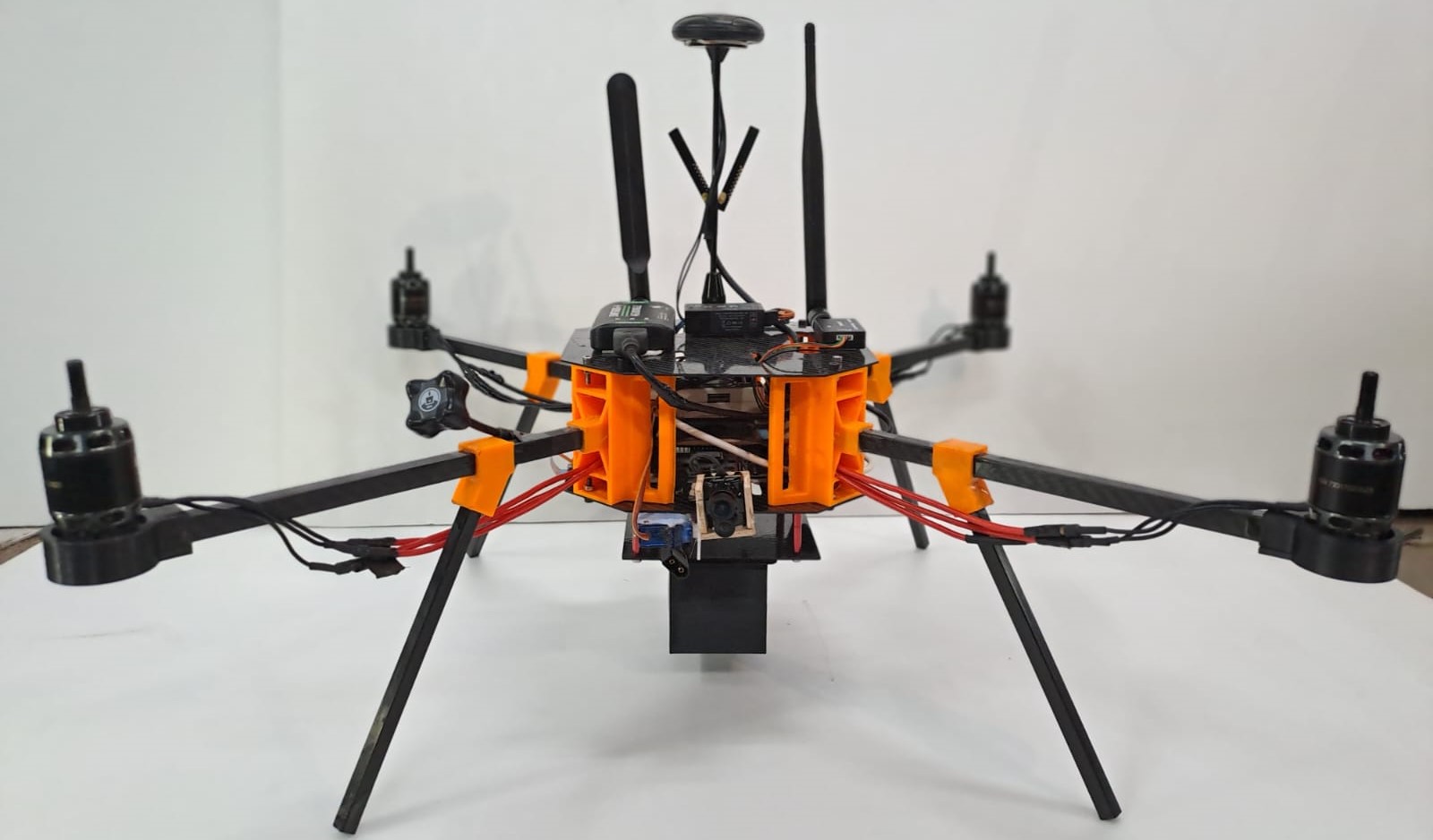} 
\caption{Final Assembled UAV}
\end{figure}

\section{Autonomous Operation}

\subsection{Autonomous Navigation Algorithm}

A list of GPS coordinates of desired waypoints is provided to the system, which is then processed by the ArduCopter autopilot. The autopilot calculates the required heading and velocity to reach the current given waypoint, and a Proportional Integral Derivative (PID) controller is employed to facilitate navigation to the specific waypoint. This GPS navigation method is employed to guide the UAV towards the vicinity of the designated payload drop location. 

\begin{figure}[h!]
\centering
\includegraphics[width=5.4in, height=2.6in]{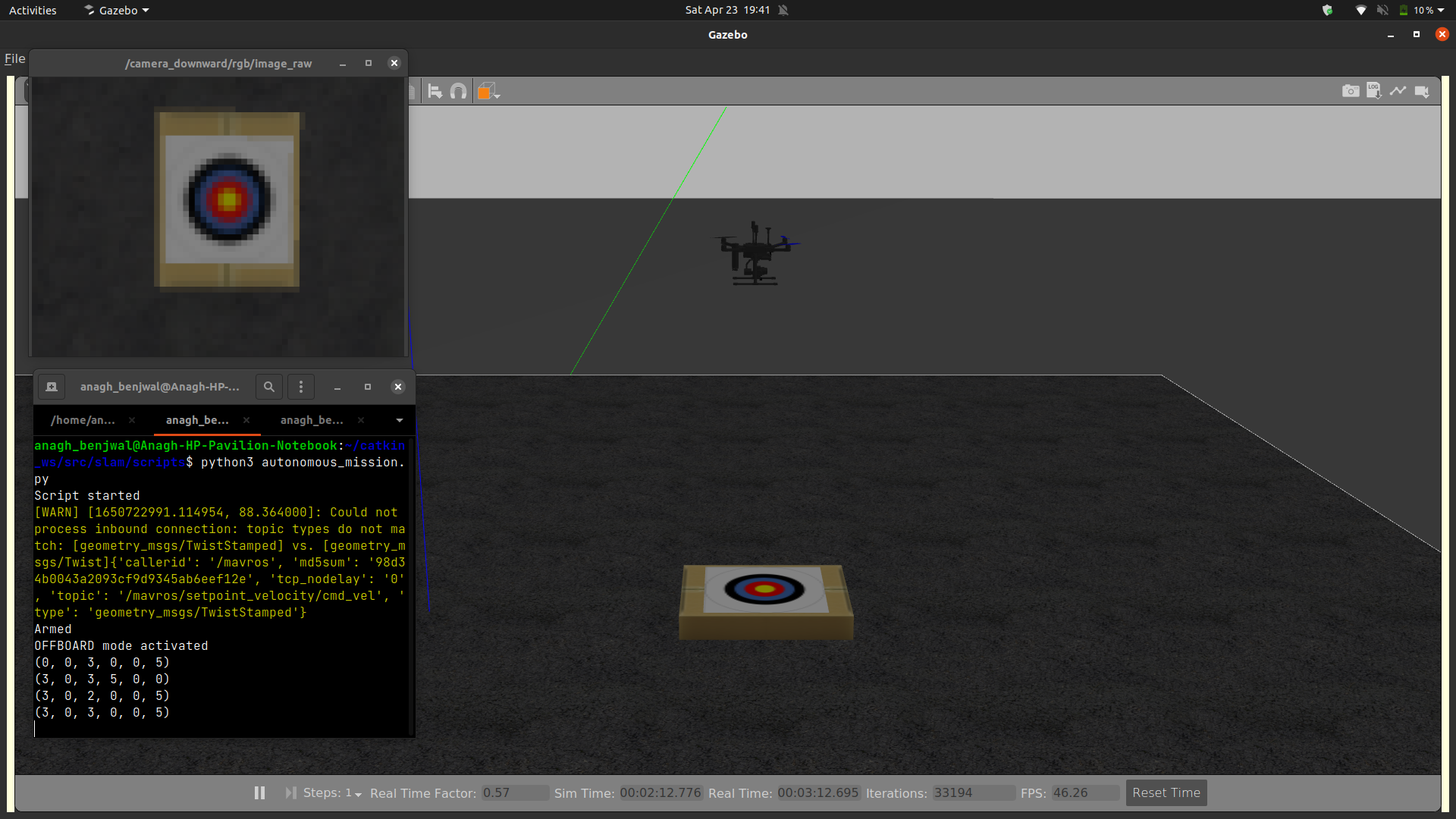} 
\caption{Simulation of Autonomous Navigation}
\end{figure}

Once the UAV is within a threshold distance in the vicinity of the payload drop location, control is transferred to the target detection algorithm. The algorithm aims to detect the target via the camera feed and then precisely align the UAV with the detected target. Upon successful alignment, the payload is dispensed from the UAV.

\begin{figure}[h!]
\centering
\includegraphics[height=5.2in, width=4.8in]{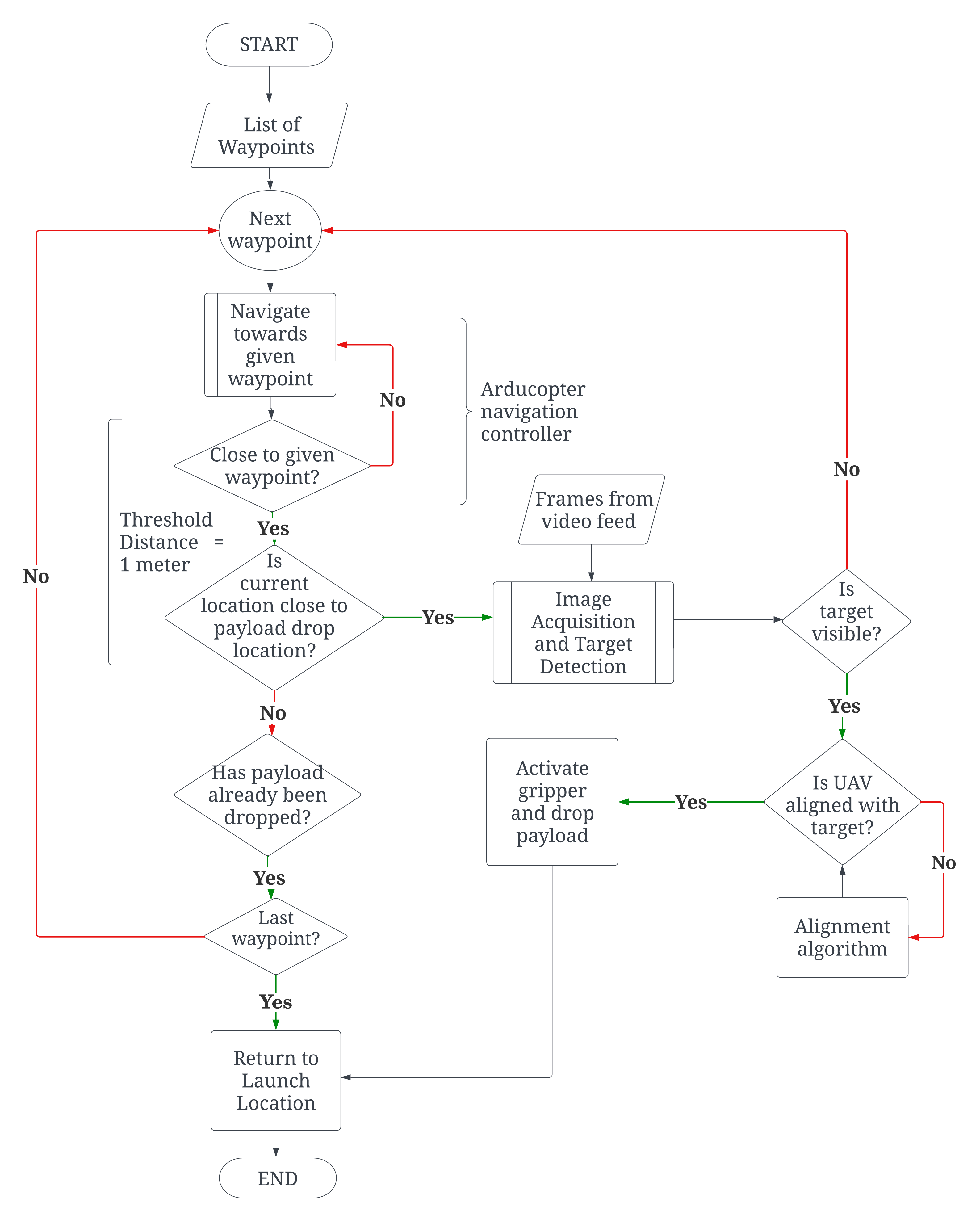} 
\caption{Logic diagram of Autonomous Operation}
\end{figure}

The autonomous navigation algorithm for the UAV was developed using the dronekit open-source Application Programming Interface (API); the autonomous capabilities of the UAV were first simulated using ArduCopter-sitl and the Gazebo simulation environment as shown in Figure 5. This was followed by extensive flight testing.

\subsection{Autonomous Identification of Target and Payload Delivery}

The use of a Convolutional Neural Network (CNN) model with EfficientNetV2 architecture \cite{tan2021efficientnetv2} was found to be most suitable for object detection after comparing various traditional computer vision and deep-learning based object detection techniques. EfficientNetv2 features adaptive regularization and compound scaling \cite{tan2021efficientnetv2}, it is ideal for the limited on-board processing power of the UAV. It is faster than traditional CNNs and achieves better accuracy than comparable lightweight models, as observed in Table 2.

\begin{figure}[h!]
\centering
\begin{minipage}{0.48\textwidth}
\includegraphics[width=2.8in]{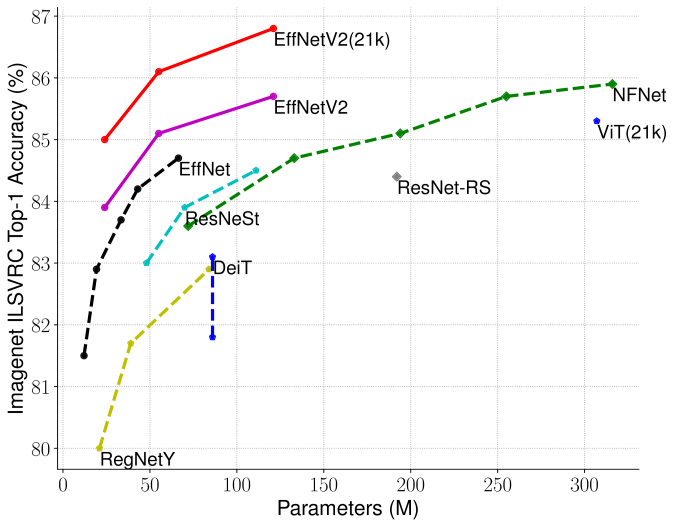} 
\end{minipage}%
\begin{minipage}{0.48\textwidth}
\includegraphics[width=2.8in]{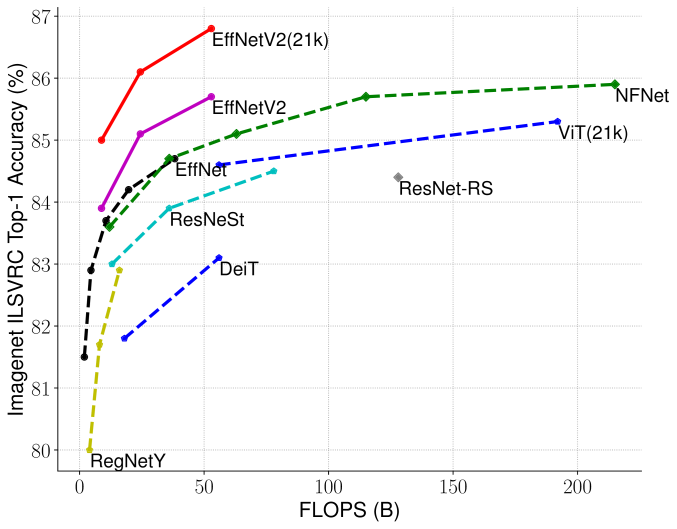} 
\end{minipage}
\begin{minipage}{0.48\textwidth}
\includegraphics[width=2.8in]{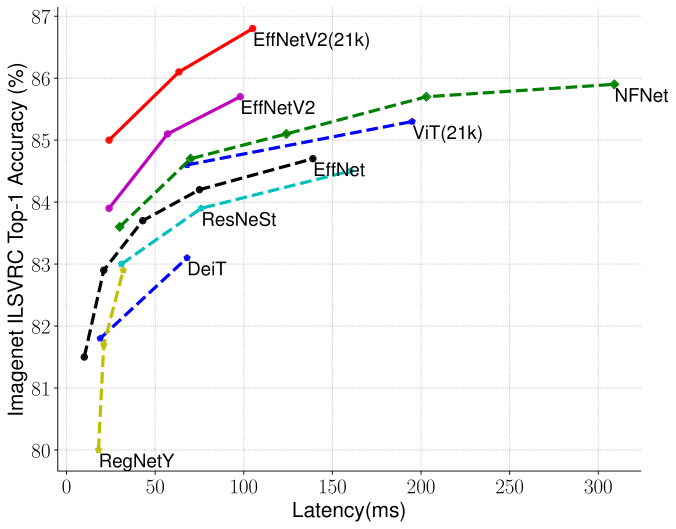} 
\end{minipage}
\begin{minipage}{0.48\textwidth}
\includegraphics[width=2.8in]{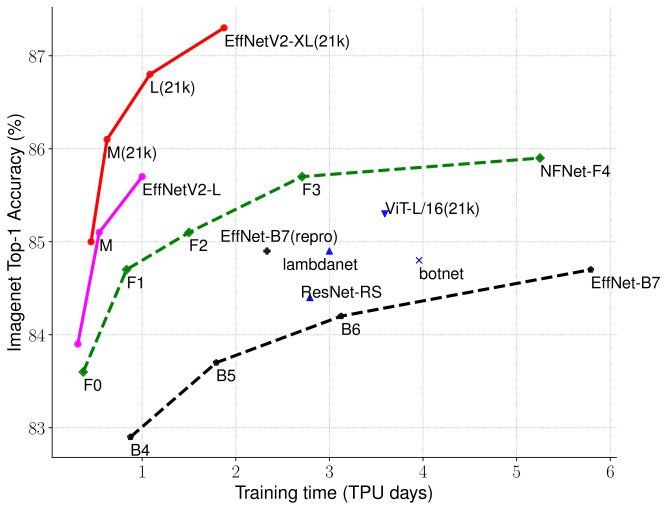} 
\end{minipage}
\caption{Comparison of Model Size, Floating Point Operations Per Second (FLOPS), Inference Latency, and training time of various model architectures against their accuracy\cite{tan2021efficientnetv2}}
\end{figure}

The model takes each frame from the video feed of the UAV's downward-facing camera as input and outputs the height, width, and pixel coordinates of the center of the bounding box as an array, this information is used to mark the center of the target in the image.

A dataset was generated by capturing aerial footage of the target from an altitude of 20-30 meters, consisting of 5 videos at 30 frames per second and 1100 keyframes annotated with a bounding box enclosing the target. The dataset includes images with targets at different angles, varied lighting and background conditions, and pictures deliberately without a target.

\begin{figure}[h!]
\centering
\begin{minipage}{0.48\textwidth}
\includegraphics[width=3in]{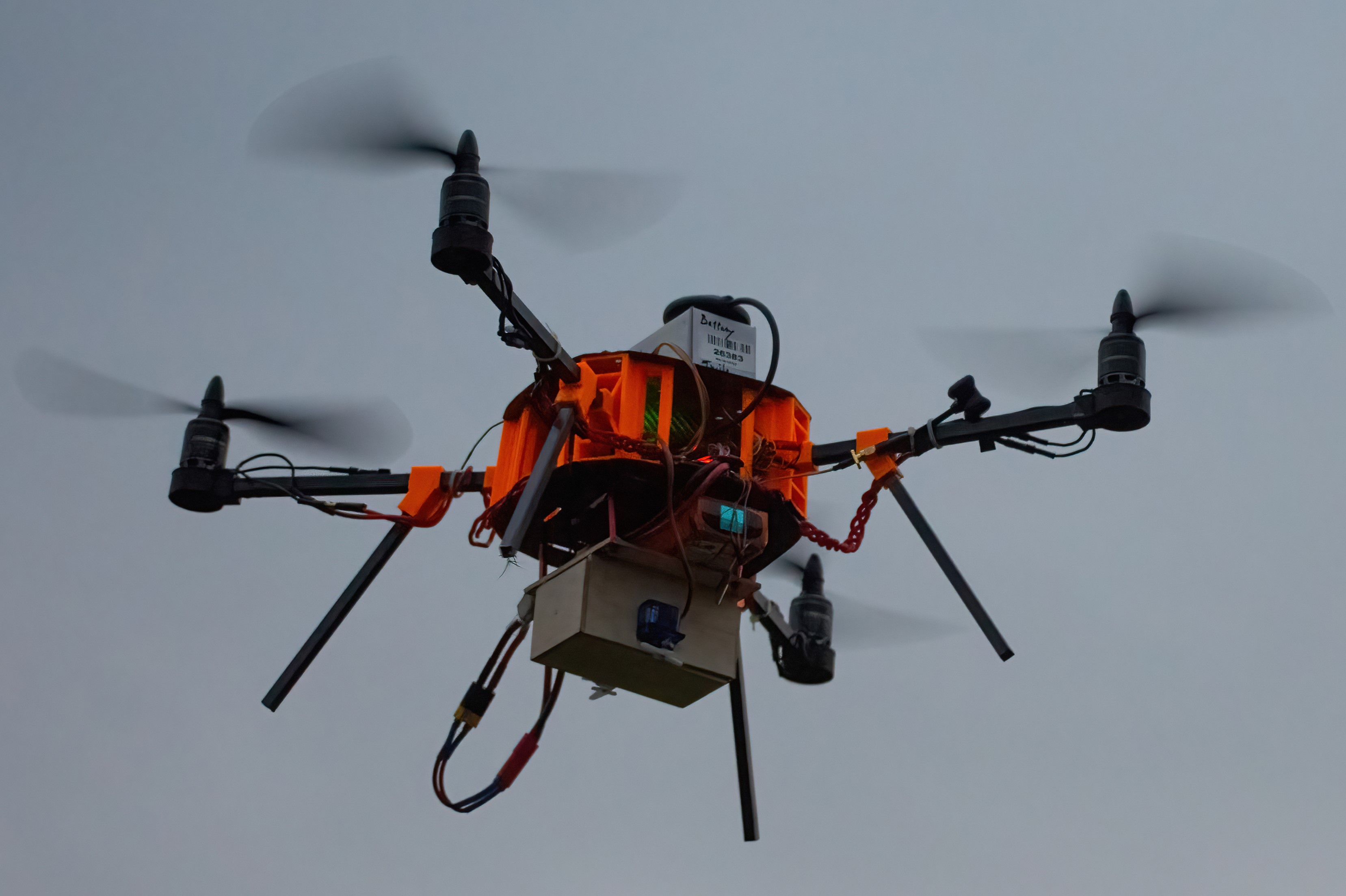}  
\caption{Flight testing of UAV}
\end{minipage}%
\begin{minipage}{0.48\textwidth}
\centering
\includegraphics[width=3in]{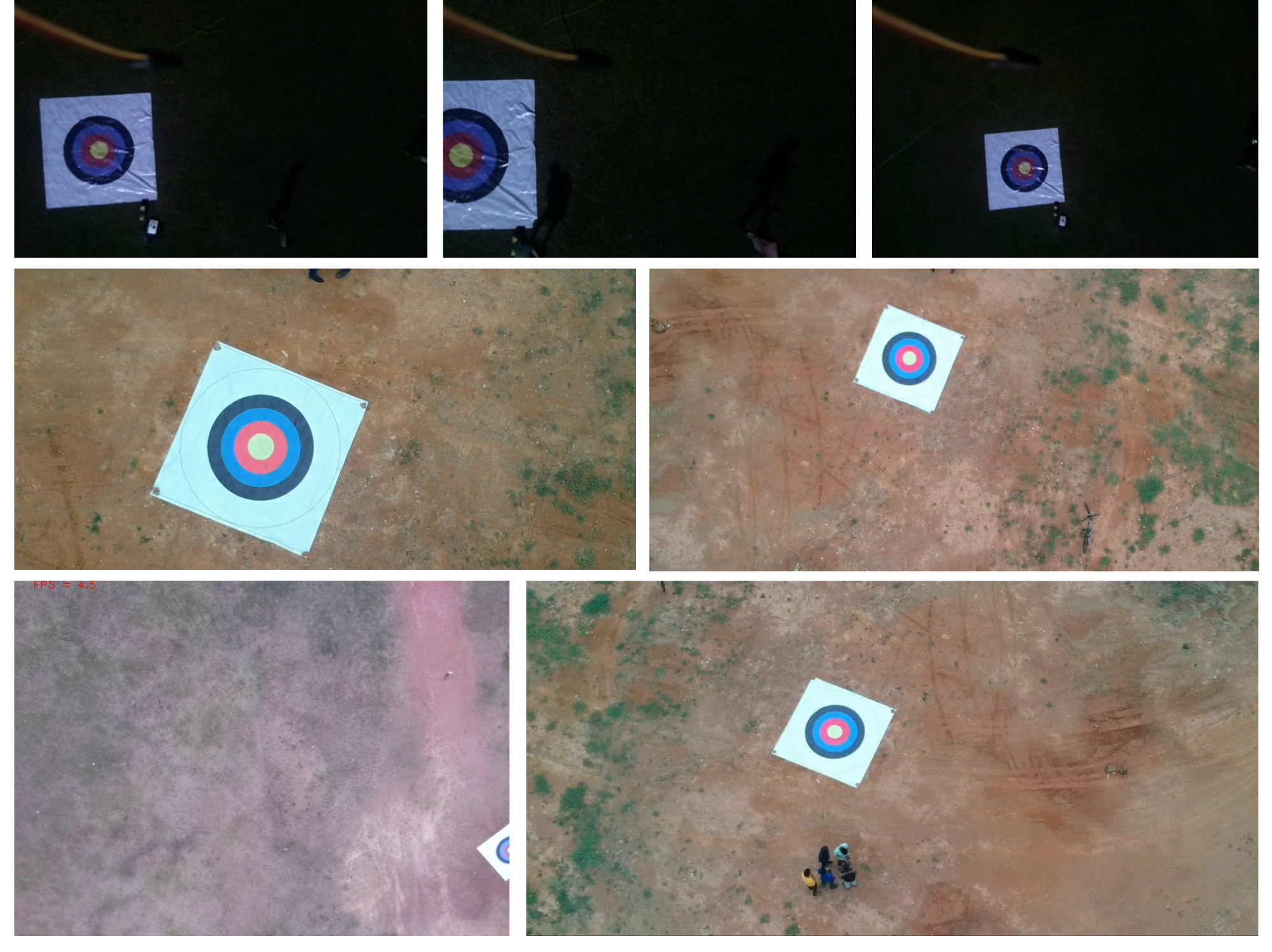} 
\caption{Samples of images from the developed dataset}
\end{minipage}

\end{figure}

A tensorflow-lite implementation of the EfficientNetv2 architecture that had been pre-trained on the ImageNet dataset was taken from  \cite{tensorflow}. The last layer of this model was re-trained for 500 epochs (keeping the other layers frozen) on the above-mentioned dataset.

\begin{table}[h!]
\caption{\label{ex}EfficientNetV2 uses 22x fewer FLOPS compared to YOLOv5-small while producing similar accuracy}
\begin{center}
\begin{tabular}{llllllll}\br
               & AP    & AP50  & AP75  & APM   & \begin{tabular}[c]{@{}l@{}}Trainable \\ Parameters\end{tabular} & GFLOPs & Accuracy \\ \mr
EfficientNetv2 & 0.476 & 0.676 & 0.676 & 0.476 & 7.4 million          & 0.7    & 96.3\%   \\
YOLOv5-small   & 0.584 & 0.742 & 0.732 & 0.696 & 7 million            & 15.9   & 99.1\% \\ \br 
\end{tabular}
\end{center}
\end{table}

To further reduce the chance of false positive predictions, a two-step verification process was employed. After a positive prediction was made by the model, the region of interest within the bounding box was extracted and subjected to color detection. If the colors of the target were present within the region of interest, the prediction was confirmed as a true positive. However, if the colors were not present, the prediction was deemed as a false positive and disregarded.

In the flight testing phase, 1500 frames were extracted from the UAV's footage. Out of these, 900 contained the target, and the model was able to accurately identify 867 of these without any false positive detections. Through extensive flight testing, the effectiveness of the proposed system was validated. The results showed that the UAV consistently achieved a horizontal positional accuracy ranging from 50 centimeters to 1 meter.

\begin{figure}[h!]
\centering
\includegraphics[width=6.4in, height=2in]{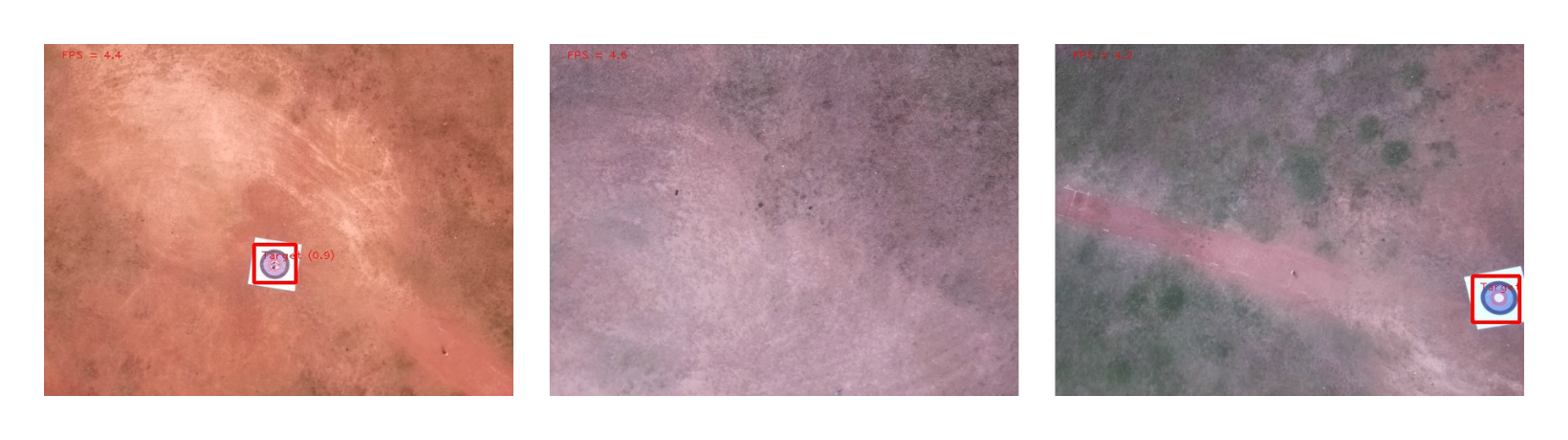}  
\caption{Examples of True Positive and True Negative predictions made by model}
\end{figure}


\section{Conclusion}

This paper details the development of a novel micro-class UAV capable of autonomous package delivery. The UAV possesses a flight time of 8 minutes, and a range of 5km while maintaining an all-up weight of 1.95kg. A two-part navigation system is proposed that utilizes GPS-based navigation to the payload drop location, followed by an object detection algorithm that identifies the marked target location in different terrains and precisely aligns the UAV with it. The object detection algorithm uses an EfficientNetv2 CNN model. A dataset was developed with images of the target in different conditions to train the model which was able to detect the target with an accuracy of 96\% despite the constrained computational power of the onboard companion computer. Extensive flight testing and simulation results show that the proposed system improves precision and provides a more robust approach to payload delivery compared to conventional GPS-only methods. The proposed system improves the average horizontal positional accuracy by up to 5 times.

\section{Future Work}

The presented autonomous payload delivery system is constrained to work only with the specific shape and form of the target that has been presented in this work. However, this algorithm can be adapted via techniques such as transfer learning to detect targets of any shape and size based on the given scenario and use case. The reliance on a specified and pre-located target area can be removed altogether by incorporating image segmentation techniques to detect areas where it is safe to drop the payload in the vicinity of the given GPS coordinates.

The accuracy of the target detection algorithm can be further improved by refining the proposed deep learning model using more advanced and modern techniques such as temporal feature networks \cite{9575941}. However, this will be heavily constrained by the onboard computational power of the UAV.

The testing of the payload delivery algorithm for this work was done from a height of 20m. In cases where the UAV will have to deliver from a lower altitude or from ground level, the presented algorithm will not work as is, and will have to be tweaked in order to facilitate the requirements of the particular scenario.

In instances where the video feed captures images at a high frame rate, there is little change in features between successive frames. In this scenario, processing each individual frame would be inefficient and unnecessary. To resolve this, a keyframe-based approach, as presented in \cite{keyframe} can be utilized to selectively process relevant frames from the video feed, saving time and computational resources.

\section*{Acknowledgments}

The authors would like to express their heartfelt gratitude to Manipal Academy of Higher Education (MAHE),  Manipal Institute of Technology, Manipal, and AeroMIT for their constant support in providing resources, funding and a productive environment. 

\section*{References}

\bibliographystyle{iopart-num}

\bibliography{AICECS}

\providecommand{\newblock}{}
\begin{thebibliography}{10}
\expandafter\ifx\csname url\endcsname\relax
  \def\url#1{{\tt #1}}\fi
\expandafter\ifx\csname urlprefix\endcsname\relax\def\urlprefix{URL }\fi
\providecommand{\eprint}[2][]{\url{#2}}

\bibitem{michelindia}
Michel A~H, Stohl R, Burke S, Kanungo B, Harpootlian A and Wille C~D {\em India and the United States: The Time Has Come to Collaborate on Commercial Drones (Sylvia Mishra)\/} (JSTOR)

\bibitem{tarasov_2022}
Tarasov K 2022 A first look at amazon's new delivery drone, slated to start deliveries this year \urlprefix\url{https://www.cnbc.com/2022/11/11/a-first-look-at-amazons-new-delivery-drone.html}

\bibitem{aurambout2019last}
Aurambout J~P, Gkoumas K and Ciuffo B 2019 {\em Last mile delivery by drones: An estimation of viable market potential and access to citizens across European cities\/} vol~11 (Springer)

\bibitem{vehiclerouting}
Dorling K, Heinrichs J, Messier G~G and Magierowski S 2016 {\em Vehicle routing problems for drone delivery\/} vol~47 (IEEE)

\bibitem{dronepsych}
Yoo W, Yu E and Jung J 2018 {\em Drone delivery: Factors affecting the public’s attitude and intention to adopt\/} vol~35 (Elsevier)

\bibitem{dronesecurity}
Altawy R and Youssef A~M 2016 {\em Security, privacy, and safety aspects of civilian drones: A survey\/} vol~1 (ACM New York, NY, USA)

\bibitem{dronedefib}
Cheskes S, McLeod S~L, Nolan M, Snobelen P, Vaillancourt C, Brooks S~C, Dainty K~N, Chan T~C and Drennan I~R 2020 {\em Improving access to automated external defibrillators in rural and remote settings: a drone delivery feasibility study\/} vol~9 (Am Heart Assoc)

\bibitem{medicinedel}
Hii M~S~Y, Courtney P and Royall P~G 2019 {\em An evaluation of the delivery of medicines using drones\/} vol~3 (MDPI)

\bibitem{productdelivery}
Bamburry D 2015 {\em Drones: Designed for product delivery\/} vol~26 (Wiley Online Library)

\bibitem{rychlicki2020analysis}
Rychlicki M, Kasprzyk Z and Rosi{\'n}ski A 2020 {\em Analysis of accuracy and reliability of different types of GPS receivers\/} vol~20 (MDPI)

\bibitem{campus2021autonomous}
Campus S~K 2021 {\em Autonomous Nano-Drone Delivery\/}

\bibitem{8324394}
Chung A~Y, Lee J~Y and Kim H 2017 {\em Autonomous mission completion system for disconnected delivery drones in urban area\/}

\bibitem{8999672}
Fanin F and Hong J~H 2019 {\em Visual Inertial Navigation for a Small UAV Using Sparse and Dense Optical Flow\/}

\bibitem{inproceedings}
Gernot C, O'Keefe K and Lachapelle G 2008 {\em Combined L1 / L2C Tracking Scheme for Weak Signal Environments\/}

\bibitem{uavcalc}
Hunsaker D and Phillips W 2013 Momentum theory with slipstream rotation applied to wind turbines

\bibitem{equationsuav}
Theys B and Schutter J~D 2020 Forward flight tests of a quadcopter unmanned aerial vehicle with various spherical body diameters vol~12 p 1756829320923565 \urlprefix\url{https://doi.org/10.1177/1756829320923565}

\bibitem{missionplanner}
Mission planner home \urlprefix\url{https://ardupilot.org/planner/}

\bibitem{raspberrypi}
Pi R Raspberry pi 4 model b \urlprefix\url{https://www.raspberrypi.com/products/raspberry-pi-4-model-b/}

\bibitem{tan2021efficientnetv2}
Tan M and Le Q 2021 {\em Efficientnetv2: Smaller models and faster training\/}

\bibitem{tensorflow}
Tensorflow Tensorflow lite \urlprefix\url{https://github.com/tensorflow/examples}

\bibitem{9575941}
Weber M, Wald T and Zöllner J~M 2021 Temporal feature networks for cnn based object detection {\em 2021 IEEE Intelligent Vehicles Symposium (IV)\/} pp 1478--1484

\bibitem{keyframe}
Girisha S, Pai M, Verma U and Pai R 2019 {\em Performance analysis of semantic segmentation algorithms for finely annotated new UAV aerial video dataset (manipaluavid)\/} vol~7 (Institute of Electrical and Electronics Engineers Inc.)

\end{thebibliography}
\end{document}